\documentclass{article}

\usepackage[utf8]{inputenc} 
\usepackage[T1]{fontenc}
\usepackage{tgtermes}
\usepackage{amssymb}
\usepackage{amsfonts}

\usepackage{tgtermes}
\usepackage{amsmath}
\usepackage{scalefnt,letltxmacro}
\LetLtxMacro{\oldtextsc}{\textsc}
\renewcommand{\textsc}[1]{\oldtextsc{\scalefont{1.10}#1}}
\usepackage[scaled=0.92]{PTSans}

\usepackage[usenames,dvipsnames]{xcolor}
\definecolor{shadecolor}{gray}{0.9}

\usepackage[parfill]{parskip}
\usepackage{afterpage}
\usepackage{framed}
\usepackage{nicefrac}

\usepackage[colorinlistoftodos,
           prependcaption,
           textsize=small,
           backgroundcolor=yellow,
           linecolor=lightgray,
           bordercolor=lightgray]{todonotes}

\usepackage{lineno}

\usepackage{ragged2e}

\DeclareRobustCommand{\parhead}[1]{\textbf{#1}~}

\usepackage{graphicx}
\usepackage[labelfont=it,labelsep=period]{caption}
\usepackage[format=hang]{subcaption}

\usepackage{booktabs}
\usepackage{arydshln} 

\usepackage{natbib}

\usepackage{algorithm}
\usepackage{algorithmic}
\usepackage{listings}
\usepackage{fancyvrb}
\fvset{fontsize=\normalsize}

\usepackage[colorlinks,linktoc=all]{hyperref}
\usepackage[all]{hypcap}
\hypersetup{citecolor=Violet}
\hypersetup{linkcolor=black}
\hypersetup{urlcolor=MidnightBlue}
\usepackage{url}

\usepackage[nameinlink]{cleveref}
\creflabelformat{equation}{#1#2#3}
\crefname{equation}{eq.}{eqs.}  
\Crefname{equation}{Eq.}{Eqs.}

\usepackage[acronym,smallcaps,nowarn]{glossaries}

\usepackage{listings}
\lstdefinestyle{alp_style}{
    commentstyle=\color{OliveGreen},
    numberstyle=\tiny\color{black!60},
    stringstyle=\color{BrickRed},
    basicstyle=\ttfamily\scriptsize,
    breakatwhitespace=false,
    breaklines=true,
    captionpos=b,
    keepspaces=true,
    numbers=none,
    numbersep=5pt,
    showspaces=false,
    showstringspaces=false,
    showtabs=false,
    tabsize=2
}
\DeclareMathOperator*{\argmax}{arg\,max}

\DeclareRobustCommand{\E}[2]{\mathbb{E}_{#1}\left[#2\right]}

\newcommand{\g}{\, | \,}
\newcommand{\prm}{\, ; \,}

\newcommand{\Bcal}{\mathcal{B}}

\newcommand{\Lcal}{\mathcal{L}}
\newcommand{\Ncal}{\mathcal{N}}
\newcommand{\Ocal}{\mathcal{O}}
\newcommand{\Scal}{\mathcal{S}}

\newacronym{ADVI}{advi}{automatic differentiation variational inference}
\newacronym{AR}{a{\small\&}r}{augment and reduce}

\newacronym{BBVI}{bbvi}{black-box variational inference}

\newacronym{CDF}{cdf}{cumulative distribution function}
\newacronym{CS-EFE}{cs-efe}{context selection for exponential family embeddings}
\newacronym{CTM}{ctm}{correlated topic model}

\newacronym[\glslongpluralkey={deep exponential families}]{DEF}{def}{deep exponential family}
\newacronym{DMIS}{dmis}{deterministic multiple importance sampling}

\newacronym{EFE}{efe}{exponential family embeddings}
\newacronym{ELBO}{elbo}{evidence lower bound}
\newacronym{EM}{em}{expectation maximization}

\newacronym{GNTS}{gn-ts}{gamma-normal time series model}
\newacronym{G-REP}{g-rep}{generalized reparameterization}

\newacronym{KL}{kl}{{K}ullback-{L}eibler}

\newacronym{LDA}{lda}{latent {D}irichlet allocation}

\newacronym{MAP}{map}{\emph{maximum a posteriori}}
\newacronym{MF}{mf}{matrix factorization}
\newacronym{MIS}{mis}{multiple importance sampling}

\newacronym{OBBVI}{o-bbvi}{overdispersed black-box variational inference}
\newacronym{OVE}{ove}{one-vs-each}

\newacronym{SVI}{svi}{stochastic variational inference}

\newacronym{VEM}{vem}{variational expectation maximization}
\newacronym{VI}{vi}{variational inference}

\usepackage[accepted]{icml2018mod}

\hypersetup{citecolor=Violet}
\hypersetup{linkcolor=black}
\hypersetup{urlcolor=MidnightBlue}

\icmltitlerunning{Augment and Reduce: Stochastic Inference for Large Categorical Distributions}

\begin{document}

\twocolumn[
\icmltitle{Augment and Reduce: Stochastic Inference for Large Categorical Distributions}


\icmlsetsymbol{equal}{*}

\begin{icmlauthorlist}
\icmlauthor{Francisco J.~R.~Ruiz}{cam,cu}
\icmlauthor{Michalis K.~Titsias}{aueb}
\icmlauthor{Adji B.~Dieng}{cu}
\icmlauthor{David M.~Blei}{cu}
\end{icmlauthorlist}

\icmlaffiliation{cam}{University of Cambridge.}
\icmlaffiliation{cu}{Columbia University.}
\icmlaffiliation{aueb}{Athens University of Economics and Business.}

\icmlcorrespondingauthor{Francisco J.~R.~Ruiz}{f.ruiz@eng.cam.ac.uk, f.ruiz@columbia.edu}

\icmlkeywords{extreme classification, large-scale classification, softmax, softmax augmentation, variational inference, stochastic variational inference, categorical distributions, discrete choice, stochastic gradient descent}

\vskip 0.3in
]



\printAffiliationsAndNotice{}  


\begin{abstract}
\vskip 0.1in
Categorical distributions are ubiquitous in machine learning, e.g., in classification, language models, and recommendation systems. However, when the number of possible outcomes is very large, using categorical distributions becomes computationally expensive, as the complexity scales linearly with the number of outcomes. To address this problem, we propose \emph{augment and reduce} (\acrshort{AR}), a method to alleviate the computational complexity. \acrshort{AR} uses two ideas: latent variable augmentation and stochastic variational inference. It maximizes a lower bound on the marginal likelihood of the data. Unlike existing methods which are specific to softmax, \acrshort{AR} is more general and is amenable to other categorical models, such as multinomial probit. On several large-scale classification problems, we show that \acrshort{AR} provides a tighter bound on the marginal likelihood and has better predictive performance than existing approaches.
\end{abstract}

\section{Introduction}
\label{sec:introduction}
\glsresetall

Categorical distributions are fundamental to many areas of machine
learning.  Examples include classification \citep{Gupta2014}, language
models \citep{bengio2006neural}, recommendation systems
\citep{Marlin2004}, reinforcement learning \citep{Sutton1998}, and neural
attention models \citep{Bahdanau2015}. They also
play an important role in discrete choice models
\citep{McFadden1978}.

A categorical is a die with $K$ sides, a discrete random variable that
takes on one of $K$ unordered outcomes; a categorical distribution gives
the probability of each possible outcome.  Categorical variables are
challenging to use when there are many possible outcomes.  Such large
categoricals appear in common applications such as image
classification with many classes, recommendation systems with many
items, and language models over large vocabularies.  In this paper, we
develop a new method for fitting and using large categorical
distributions.

The most common way to form a categorical is through the softmax
transformation, which maps a $K$-vector of reals to a distribution of
$K$ outcomes.  Let $\psi$ be a real-valued $K$-vector. The softmax
transformation is
\begin{align}
  \label{eq:softmax}
  p(y = k \g \psi)
  =
  \frac{\exp\left\{\psi_{k}\right\}}
  {\sum_{k'} \exp\left\{\psi_{k'}\right\}}.
\end{align}
Note the softmax is not the only way to map real vectors to
categorical distributions; for example, the multinomial probit \citep{Albert1993}
is an alternative.  Also note that in many applications, such as in
multiclass classification, the parameter $\psi_k$ is a function of
per-sample features $x$.  For example, a linear classifier forms a
categorical over classes through a linear combination,
$\psi_{k} = w_k^\top x$. 



We usually fit a categorical with maximum likelihood estimation or
any other closely related strategy.  Given a dataset $y_{1:N}$ of
categorical data---each $y_n$ is one of $K$ values---we aim to
maximize the log likelihood,
\begin{align}
  \label{eq:likelihood}
  \Lcal_{\textrm{log likelihood}} = \sum_{n=1}^{N} \log p(y_n \g \psi).
\end{align}
Fitting this objective requires evaluating both the log probability and
its gradient.

\Cref{eq:softmax,eq:likelihood} reveal the challenge to
using large categoricals.  Evaluating the log probability and
evaluating its gradient are both $\Ocal(K)$ operations.  But this is not
OK: most algorithms for fitting categoricals---for example, stochastic
gradient ascent---require repeated evaluations of both gradients and
probabilities.  When $K$ is large, these algorithms are
prohibitively expensive.

Here we develop a method for fitting large categorical
distributions, including the softmax but also more generally.  It
is called \gls{AR}.  \gls{AR} rewrites the categorical
distribution with an auxiliary variable $\varepsilon$,
\begin{align}
  \label{eq:auxillary}
  p(y \g \psi) = \int p(y, \varepsilon \g \psi) d\varepsilon.
\end{align}
\gls{AR} then replaces the expensive log probability
with a variational bound on the integral in
\Cref{eq:auxillary}.  Using stochastic variational
methods \citep{Hoffman2013}, the cost to evaluate the bound
(or its gradient) is far below $\Ocal(K)$.

Because it relies on variational methods, \gls{AR} provides a lower
bound on the marginal likelihood of the data. With this bound, we
can embed \gls{AR} in a larger algorithm for fitting a categorical,
e.g., a (stochastic) \gls{VEM} algorithm \citep{Beal2003}.
Though we focus on maximum likelihood, we can also use \gls{AR} in other
algorithms that require $\log p(y \g \psi)$ or its gradient, e.g., fully
Bayesian approaches \citep{Gelman2003}
or the \textsc{reinforce} algorithm \citep{Williams1992}.


We study \gls{AR} on linear classification tasks with up to $10^4$ classes.
On simulated and real data, we find that it
provides accurate estimates of the categorical probabilities and gives
better performance than existing approaches.


\parhead{Related work.}  There are many methods to reduce the
cost of large categorical distributions, particularly
under the softmax transformation. These include methods that
approximate the exact computations
\citep{Gopal2013,Vijayanarasimhan2014}, those that rely on sampling
\citep{Bengio2003quick,Mikolov2013distributed,Devlin2014,Ji2016,Botev2017},
those that use approximations and distributed computing
\citep{Grave2017}, double-sum formulations \citep{Raman2017,Fagan2018},
and those that avail themselves of other techniques
such as noise contrastive estimation \citep{Smith2005,Gutmann2010} or
random nearest neighbor search \citep{Mussmann2017}.




Other methods change the model.  They might replace the
softmax transformation with a hierarchical or stick-breaking model
\citep{Kurynski1988,Morin2005,Tsoumakas2008,Beygelzimer2009,Dembczynski2010,Khan2012}.
These approaches can be successful, but the structure of the hierarchy may influence
the learned probabilities.
Other methods replace the softmax with a scalable spherical family of losses
\citep{Vincent2015,deBrebisson2016}.




\gls{AR} is different from all of these techniques. Unlike many of them, it
provides a lower bound on the log probability rather than an
approximation.  The bound is useful because it can naturally be embedded in
algorithms like stochastic \gls{VEM}.  Further, the \gls{AR}
methodology applies to transformations beyond the softmax.  In this paper, we
study large categoricals via softmax, multinomial probit, and
multinomial logistic.  \gls{AR} is the first scalable 
approach for the two latter models.  It accelerates any transformation that
can be recast as an additive noise model
\citep[e.g.,][]{Gumbel1954,Albert1993}.


The approach that most closely relates to \gls{AR} is the \gls{OVE}
bound of \citet{Titsias2016}, which is a lower bound of the softmax.
Like the other related methods, it is narrower than
\gls{AR} in that it does not apply to transformations beyond the softmax.
We also empirically compare \gls{AR} to \gls{OVE} in
\Cref{sec:experiments}.  \gls{AR} provides a tighter lower
bound and yields better predictive performance.


\section{Augment and Reduce}
\label{sec:method}
\glsresetall

We develop \gls{AR}, a method for computing with large categorical
random variables.

\parhead{The utility perspective.} \gls{AR} uses the additive noise
model perspective on the categorical, which we refer to as the utility
perspective.  Define a \emph{mean utility} $\psi_k$ for each possible outcome
$k\in\{1,\ldots,K\}$.  To draw a variable $y$ from a categorical, we
draw a zero-mean noise term $\varepsilon_k$ for each possible outcome
and then choose the value that maximizes the \emph{realized utility}
$\psi_k + \varepsilon_k$.  This corresponds to the following process,
\begin{align}
  \begin{split}
    \label{eq:max_utility}
    \varepsilon_k &\sim \phi(\cdot), \quad k \in \{1, \ldots, K\}, \\
    y &= \argmax_k\left(\psi_k + \varepsilon_k \right).
  \end{split}
\end{align}
Note the errors $\varepsilon_k$ are drawn fresh each time we draw a variable $y$.  
We assume that the errors are independent of each other, independent of the mean utility
$\psi_k$, and identically distributed according to some distribution
$\phi(\cdot)$.

Now consider the model where we marginalize the errors from \Cref{eq:max_utility}.
This results in a distribution $p(y \g \psi)$, a categorical that transforms
$\psi$ to the simplex.
Depending on the distribution of the errors, this
induces different transformations.
For example, a standard Gumbel distribution recovers the
softmax transformation; a standard Gaussian recovers the multinomial
probit transformation; a standard logistic recovers the multinomial
logistic transformation.



Typically, the mean utility $\psi_k$ is a function of observed
features $x$, e.g., $\psi_k=x^\top w_k$ in linear models or
$\psi_k=f_{w_k}(x)$ in non-linear settings.  In both cases, $w_k$
are model parameters, relating the features to mean utilities.

Let us focus momentarily on a linear classification problem under the
softmax model. For each observation $n$, the mean utilities are
$\psi_{nk}=x_n^\top w_k$ and the random errors $\varepsilon_{nk}$ are
Gumbel distributed. After marginalizing out the errors, the
probability that observation $n$ is in class $k$ is given by \Cref{eq:softmax},
$p(y_n=k \g x_n, w) \propto \exp\{ x_n^\top w_k \}$.
Fitting the classifier involves learning the weights $w_k$ that
parameterize $\psi$.  For example, maximum likelihood 
uses gradient ascent to maximize $\sum_n \log p(y_n \g x_n,w)$ with
respect to $w$.


\parhead{Large categoricals.}  When the number of outcomes $K$ is
large, the normalizing constant of the softmax is a
computational burden; it is $\Ocal(K)$.  Consequently, it is
burdensome to calculate useful quantities like
$\log p(y_n \g x_n,w)$ and its gradient
$\nabla_w \log p(y_n \g x_n,w)$.  As an ultimate consequence,
maximum likelihood estimation is slow---it needs to evaluate the
gradient for each $n$ at each iteration.

Its difficulty scaling is not unique to the softmax.
Similar issues arise for the multinomial probit and multinomial
logistic.  With these transformations as well, evaluating likelihoods
and related quantities is $\Ocal(K)$.

\subsection{Augment and reduce}

We introduce \gls{AR} to relieve this burden.  \gls{AR} accelerates
training in models with categorical distributions and a large number
of outcomes.

Rather than operating directly on the marginal $p(y \g \psi)$, \gls{AR}
\textit{augments} the model with one of the error terms and forms a
joint $p(y,\varepsilon\g \psi)$. (We drop the subscript $n$ to avoid
cluttered notation.)  This augmented model has a desirable property:
its log-joint is a sum over all the possible outcomes.  \gls{AR} then
\textit{reduces}---it subsamples a subset of outcomes to construct
estimates of the log-joint and its gradient. As a result, its complexity
relates to the size of the subsample, not the total number of
outcomes $K$.

\parhead{The augmented model.} Let $\phi(\varepsilon)$ be the
distribution over the error terms, and
$\Phi(\varepsilon)=\int_{-\infty}^\varepsilon \phi(\tau)d\tau$ the
corresponding \gls{CDF}. The marginal
probability of outcome $k$ is the probability that its realized utility
($\psi_k + \varepsilon_k$) is greater than all others,
\begin{align*}
  p(y=k\g \psi) &= \textrm{Pr}\left(\psi_k + \varepsilon_k \geq
                  \psi_{k^\prime} + \varepsilon_{k^\prime}\;\; \forall
                  k^\prime \neq k \right).
\end{align*}
We write this probability as an integral over the $k$th error
$\varepsilon_k$ using the \gls{CDF} of the other errors,
\begin{align}
    p(y=k\g \psi) &= \!
    \int_{-\infty}^{+\infty} \!\!\!\!\! \phi(\varepsilon_k)\! \Bigg( \!
      \prod_{k^\prime\neq k} \!
      \int_{-\infty}^{\varepsilon_{k}+\psi_k-\psi_{k^\prime}}
      \!\!\!\!\!\phi(\varepsilon_{k^\prime}) d\varepsilon_{k^\prime} \!\!\Bigg)
    \! d\varepsilon_k  \nonumber \\ 
    & = \! \int_{-\infty}^{+\infty} \!\!\!\!\! \phi(\varepsilon) \! \Bigg( \!
      \prod_{k^\prime\neq k} \! \Phi(\varepsilon +
      \psi_{k}-\psi_{k^\prime})\! \Bigg)\! d\varepsilon.
      \label{eq:girolami_rogers}
\end{align}
(We renamed the dummy variable $\varepsilon_k$ as $\varepsilon$
to avoid clutter.) \Cref{eq:girolami_rogers} is the
same as found by \citet{Girolami2006} for the multinomial probit
model, although we do not assume a Gaussian density
$\phi(\varepsilon)$. Rather, we only assume that we can evaluate both
$\phi(\varepsilon)$ and $\Phi(\varepsilon)$.

We derived \Cref{eq:girolami_rogers} from the utility perspective,
which encompasses many common models. We obtain the softmax by choosing a
standard Gumbel distribution for $\phi(\varepsilon)$, in which case
\Cref{eq:girolami_rogers,eq:softmax} are equivalent. We
obtain the multinomial probit by choosing a standard Gaussian
distribution over the errors, and in this case the integral in
\Cref{eq:girolami_rogers} does not have a closed form. Similarly, we
obtain the multinomial logistic by choosing a standard logistic
distribution $\phi(\varepsilon)$.  
What is important is that regardless of the model, the cost
to compute the marginal probability $p(y=k\g \psi)$ is $\Ocal(K)$.

\begin{figure}[t]
  \centering
  \begin{subfigure}[b]{0.3\columnwidth}
    \centering
    \includegraphics[height=40pt]{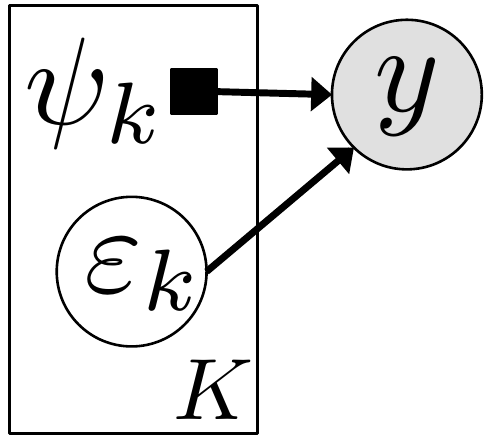}
    \caption{}
  \end{subfigure}%
  \begin{subfigure}[b]{0.3\columnwidth}
    \centering
    \includegraphics[height=40pt]{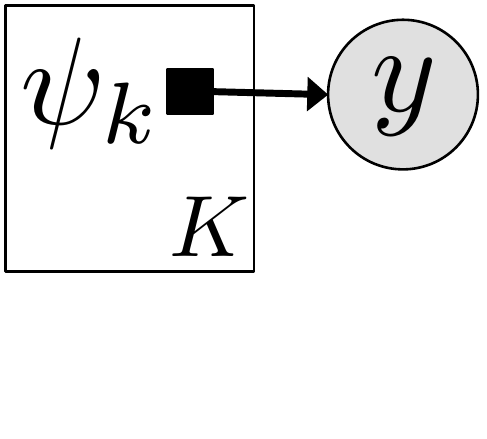}
    \caption{}
  \end{subfigure}%
  \begin{subfigure}[b]{0.3\columnwidth}
    \centering
    \includegraphics[height=40pt]{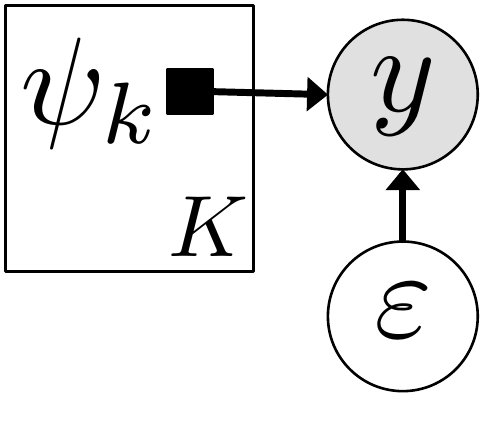}
    \caption{}
  \end{subfigure}
  \vspace*{-10pt}
  \caption{\textit{(a)} Illustration of the parameterization of a categorical model in terms of the utilities $\psi_k+\varepsilon_k$, where $\psi_k$ is the mean utility and $\varepsilon_k$ is an error term. The observed outcome is $y=\argmax_k(\psi_k+\varepsilon_k)$. \textit{(b)} In this model, the error terms have been marginalized out. This is the most common model for categorical distributions; it includes the softmax and multinomial probit. \textit{(c)} The augmented model that we consider for \acrshort{AR}. All error terms have been integrated out, except one. In this model, the log-joint involves a summation over the possible outcomes $k$, enabling fast unbiased estimates of the log probability and its gradient.\label{fig:models}}
  \vspace*{-7pt}
\end{figure}

We now augment the model with the auxiliary latent variable
$\varepsilon$ to form the joint distribution
$p(y, \varepsilon \g \psi)$,
\begin{align}
  \label{eq:joint_augmented}
  p(y=k, \varepsilon \g \psi) = \phi(\varepsilon) \prod_{k^\prime\neq
  k} \Phi(\varepsilon + \psi_{k}-\psi_{k^\prime}).
\end{align}
This is a model that includes the $k$th error term from
\Cref{eq:max_utility} but marginalizes out all the other errors.  By
construction, marginalizing $\varepsilon$ from
\Cref{eq:joint_augmented} recovers the original model $p(y\g \psi)$
in \Cref{eq:girolami_rogers}. \Cref{fig:models} illustrates this idea.

\citet{Riihimaki2013} used \Cref{eq:joint_augmented} in the nested
expectation propagation for Gaussian process classification.
We use it to scale learning with categorical distributions.


\parhead{The variational bound.}  
The augmented model in \Cref{eq:joint_augmented} involves one latent variable $\varepsilon$.
But our goal is to calculate the marginal $\log p(y\g \psi)$ and its gradient. 
\gls{AR} derives a variational lower bound on $\log p(y\g \psi)$ using the joint in \Cref{eq:joint_augmented}.
Define $q(\varepsilon)$ to be a variational distribution on the auxiliary variable. The bound is
$\log p(y\g \psi)\geq \Lcal$, where
\begin{align}
    \Lcal & = \mathbb{E}_{q(\varepsilon)} \! \Big[ \log p(y=k, \varepsilon \g \psi) - \log q(\varepsilon) \Big]\! \label{eq:elbo} \\
    & = \mathbb{E}_{q(\varepsilon)} \! \Big[ \log\phi(\varepsilon) + \!\! \sum_{k^\prime\neq k}\! \log\Phi(\varepsilon + \psi_{k}-\psi_{k^\prime}) - \log q(\varepsilon) \Big]\!. \nonumber
\end{align}
In \Cref{eq:elbo}, $\Lcal$ is the \gls{ELBO}; it is tight when $q(\varepsilon)$ is equal to the posterior of $\varepsilon$ given $y$, $p(\varepsilon\g y, \psi)$ \citep{Jordan1999,Blei2016}.

The \gls{ELBO} contains a summation over the outcomes $k^\prime \neq k$. \gls{AR} exploits this property to reduce complexity, as we describe below. Next we show how to use the bound in a \gls{VEM} procedure and we describe the reduce step of \gls{AR}. 

\parhead{Variational expectation maximization.}
Consider again a linear classification task, where we have a dataset of features $x_n$ and labels $y_n\in\{1,\ldots,K\}$ for $n=1,\ldots,N$. The mean utility for each observation $n$ is $\psi_{nk}=w_k^\top x_n$, and the goal is to learn the weights $w_k$ by maximizing the log likelihood $\sum_n \log p(y_n \g x_n,w)$.

\gls{AR} replaces each term in the data log likelihood with its bound using \Cref{eq:elbo}. The objective becomes $\sum_n \Lcal^{(n)}$. Maximizing this objective requires an iterative process with two steps. In one step, \gls{AR} optimizes the objective with respect to $w$. In the other step, \gls{AR} optimizes each $\Lcal^{(n)}$ with respect to the variational distribution. The resulting procedure takes the form of a \gls{VEM} algorithm \citep{Beal2003}.

The \gls{VEM} algorithm requires optimizing the \gls{ELBO} with respect to $w$ and the variational
distributions.\footnote{Note that maximizing the \gls{ELBO} in \Cref{eq:elbo} with respect to the distribution $q(\varepsilon)$ is equivalent to minimizing the Kullback-Leibler divergence from $q(\varepsilon)$ to the posterior $p(\varepsilon\g y, \psi)$.}
This is challenging for two reasons. First, the expectations in \Cref{eq:elbo} might not be tractable. Second, the cost to compute the gradients of \Cref{eq:elbo} is still $\Ocal(K)$. 

\Cref{sec:algorithm} addresses these issues. To sidestep the intractable expectations, \gls{AR} forms unbiased Monte Carlo estimates of the gradient of the \gls{ELBO}. To alleviate the computational complexity, \gls{AR} uses stochastic optimization, subsampling a set of outcomes $k^\prime$.


\parhead{Reduce by subsampling.}  
The subsampling step in the \gls{VEM} procedure
is one of the key ideas behind \gls{AR}.
Since \Cref{eq:elbo} contains a summation over the outcomes
$k^\prime\neq k$, we can apply stochastic optimization techniques to obtain
unbiased estimates of the \gls{ELBO} and its gradient.

More specifically, consider the gradient of the \gls{ELBO} in \Cref{eq:elbo} with respect to $w$
(the parameters of $\psi$). It is
\begin{equation*}
  \nabla_{w} \Lcal = 
  \sum_{k^\prime\neq k} \mathbb{E}_{q(\varepsilon)} \big[ \nabla_{w} \log\Phi(\varepsilon + \psi_{k}-\psi_{k^\prime}) \big].
\end{equation*}
\gls{AR} estimates this by first randomly sampling a subset of outcomes
$\Scal\subseteq\{1,\ldots,K\}\diagdown\{k\}$ of size $|\Scal|$.
\gls{AR} then uses the outcomes in $\Scal$ to approximate the gradient,
\begin{equation*}
  \widetilde{\nabla}_{w} \Lcal = 
  \frac{K-1}{|\Scal|}
  \sum_{k^\prime\in \Scal}  \mathbb{E}_{q(\varepsilon)} \big[\nabla_{w} \log\Phi(\varepsilon + \psi_{k}-\psi_{k^\prime}) \big].
\end{equation*}
This is an unbiased estimator\footnote{This is not the only way to construct an unbiased estimator. Alternatively, we can draw the outcomes $k^\prime$ using importance sampling, taking into account the frequency of each class. We leave this for future work.}
of the gradient $\nabla_{w} \Lcal$. Crucially, \gls{AR} only needs to iterate over $|\Scal|$ outcomes to obtain it, reducing the complexity to $\Ocal(|\Scal|)$.

The reduce step is also applicable to optimize the \gls{ELBO} with respect to $q(\varepsilon)$.
\Cref{sec:algorithm} gives further details about the stochastic \gls{VEM} procedure in different settings.

\section{Algorithm Description}
\label{sec:algorithm}
\glsresetall

Here we provide the details to run the \gls{VEM} algorithm for the softmax model (\Cref{sec:softmax_augment}) and for more general models including the multinomial probit and multinomial logistic (\Cref{sec:other_augment}). These models only differ in the prior over the errors $\phi(\varepsilon)$.

\Gls{AR} is not limited to point-mass estimation of the parameters $w$. It is straightforward to extend the algorithm to perform posterior inference on $w$ via stochastic variational inference, but for simplicity we describe maximum likelihood estimation.

\subsection{Augment and Reduce for Softmax}\label{sec:softmax_augment}
In the softmax model, the distribution over the error terms is a standard Gumbel \citep{Gumbel1954},
\begin{equation*}
	\phi_{\textrm{softmax}}(\varepsilon) = \exp\{ -\varepsilon-e^{-\varepsilon} \},
	\;\; \Phi_{\textrm{softmax}}(\varepsilon)=\exp\{-e^{-\varepsilon}\}.
\end{equation*}
In this model, the optimal distribution $q^{\star}(\varepsilon)$, which achieves equality in the bound, has closed-form expression:
\begin{equation*}
	q^{\star}_{\textrm{softmax}}(\varepsilon) = \textrm{Gumbel}(\varepsilon\prm\log\eta^{\star},1),
\end{equation*}
with $\eta^{\star} =  1+ \sum_{k^\prime \neq k} e^{\psi_{k^\prime}-\psi_k}$. However, even though $q_{\textrm{softmax}}^\star(\varepsilon)$ has an analytic form, its parameter $\eta^{\star}$ is computationally expensive to obtain because it involves a summation over $K-1$ classes. Instead, we set
\begin{equation*}
	q_{\textrm{softmax}}(\varepsilon\prm \eta)= \textrm{Gumbel}(\varepsilon\prm\log\eta,1).
\end{equation*}
Substituting this choice for $q_{\textrm{softmax}}(\varepsilon\prm \eta)$ into \Cref{eq:elbo} gives the following \gls{ELBO}:
\begin{equation}\label{eq:elbo_softmax}
	\Lcal_{\textrm{softmax}} = 1 - \log(\eta) - \frac{1}{\eta}\left(1+\sum_{k^\prime \neq k} e^{\psi_{k^\prime}-\psi_k}\right).
\end{equation}
\Cref{eq:elbo_softmax} coincides with the log-concavity bound \citep{Bouchard2007,Blei2007}, although we have derived it from a completely different perspective. This derivation allows us to optimize $\eta$ efficiently, as we describe next.

The $\textrm{Gumbel}(\varepsilon\prm \log\eta,1)$ is an exponential family distribution whose natural parameter is $\eta$. This allows us to use natural gradients in the stochastic inference procedure. \gls{AR} iterates between a local step, in which we update $\eta$, and a global step, in which we update the parameters $\psi$.

In the local step (\textsc{e} step), we optimize $\eta$ by taking a step in the direction of the noisy natural gradient, yielding $\eta^\textrm{new}=(1-\alpha)\eta^\textrm{old}+\alpha\widetilde{\eta}$. Here, $\widetilde{\eta}$ is an estimate of the optimal natural parameter, which we obtain using a random set of outcomes, i.e., $\widetilde{\eta} =  1+ \frac{K-1}{|\Scal|}\sum_{k^\prime \in \Scal} e^{\psi_{k^\prime}-\psi_k}$, where $\Scal\subseteq\{1,\ldots,K\}\diagdown\{k\}$.
The parameter $\alpha$ is the step size; it must satisfy the Robbins-Monro conditions \citep{Robbins1951,Hoffman2013}.

In the global step (\textsc{m} step), we take a gradient step with respect to $w$ (the parameters of $\psi$), holding $\eta$ fixed. Similarly, we can estimate the gradient of \Cref{eq:elbo_softmax}
with complexity $\Ocal(|\Scal|)$ by leveraging stochastic optimization.

\Cref{alg:augmented_sm_vi} summarizes the procedure for a classification task. In this example, the dataset consists of $N$ datapoints $(x_n,y_n)$, where $x_n$ is a feature vector and $y_n\in\{1,\ldots,K\}$ is the class label. Each observation is associated with its parameters $\psi_{nk}$; e.g., $\psi_{nk}=x_n^\top w_k$. We posit a softmax likelihood, and we wish to infer the weights via maximum likelihood using $\gls{AR}$. Thus, the objective function is $\sum_n\Lcal_{\textrm{softmax}}^{(n)}$. (It is straightforward to obtain the maximum a posteriori solution by adding a regularizer.) At each iteration, we process a random subset of observations as well as a random subset of classes for each one.

Finally, note that we can perform posterior inference on the parameters $w$ (instead of maximum likelihood) using \gls{AR}. One way is to consider a variational distribution $q(w)$ and take gradient steps with respect to the variational parameters of $q(w)$ in the global step, using the reparameterization trick \citep{Rezende2014,Titsias2014_doubly,Kingma2014} to approximate that gradient.
In the local step, we only need to evaluate the moment generating function, estimating the optimal natural parameter as $\widetilde{\eta} =  1+ \frac{K-1}{|\Scal|}\sum_{k^\prime \in \Scal} \E{q(w)}{e^{\psi_{k^\prime}-\psi_k}}$.




\newlength{\textfloatsepsave}
\setlength{\textfloatsepsave}{\textfloatsep}
\setlength{\textfloatsep}{0.10in} 

\begin{algorithm}[tb]
	\caption{Softmax \acrshort{AR} for classification}
	\label{alg:augmented_sm_vi}
	\begin{algorithmic}
		\STATE {\bfseries Input:} data $(x_n,y_n)$, minibatch sizes $|\Bcal|$ and $|\Scal|$
		\STATE {\bfseries Output:} weights $w=\{w_k\}_{k=1}^K$
		\STATE Initialize all weights and natural parameters
		\FOR{iteration $t=1,2,\ldots,$}
			\STATE \verb|# Sample minibatches:|
			\STATE Sample a minibatch of data, $\Bcal\subseteq\{1,\ldots,N\}$
			\FOR{$n\in\Bcal$}
				\STATE Sample a set of labels, $\Scal_n\subseteq\{1,\ldots,K\}\diagdown\{y_n\}$
			\ENDFOR
			\STATE \verb|# Local step (E step):|
			\FOR{$n\in\Bcal$}
				\STATE Compute $\widetilde{\eta}_n =  1+ \frac{K-1}{|\Scal|}\sum_{k^\prime \in \Scal_n} e^{\psi_{nk^\prime}-\psi_{ny_n}}$
				\STATE Update natural param., $\eta_n \leftarrow (1-\alpha^{(t)})\eta_n + \alpha^{(t)} \widetilde{\eta}_n$
			\ENDFOR
			\STATE \verb|# Global step (M step):|
			\STATE Set $g = - \frac{N}{|\Bcal|}\frac{K-1}{|\Scal|}\sum_{n\in \Bcal} \frac{1}{\eta_n} \sum_{k^\prime\in\Scal_n}\!\! \nabla_{w} e^{\psi_{nk^\prime}-\psi_{ny_n}}$
			\STATE Gradient step on the weights, $w \leftarrow w + \rho^{(t)}g$
		\ENDFOR
	\end{algorithmic}
\end{algorithm}

\setlength{\textfloatsep}{\textfloatsepsave}

\subsection{Augment and Reduce for Other Models}\label{sec:other_augment}

For most models, the expectations of the \gls{ELBO} in \Cref{eq:elbo} are intractable, and there is no closed-form solution for the optimal variational distribution $q^\star(\varepsilon)$. Fortunately, we can apply \gls{AR}, using the reparameterization trick to build Monte Carlo estimates of the gradient of the \gls{ELBO} with respect to the variational parameters \citep{Rezende2014,Titsias2014_doubly,Kingma2014}.

More in detail, consider the variational distribution $q(\varepsilon\prm\nu)$, parameterized by some variational parameters $\nu$. We assume that this distribution is reparameterizable, i.e., we can sample from $q(\varepsilon\prm\nu)$ by first sampling an auxiliary variable $u\sim q^{\textrm{(rep)}}(u)$ and then setting $\varepsilon=T(u\prm\nu)$.

In the local step, we fit $q(\varepsilon\prm\nu)$ by taking a gradient step of the \gls{ELBO} with respect to the variational parameters $\nu$. 
Since the expectations in \Cref{eq:elbo} are not tractable, we obtain Monte Carlo estimates by sampling $\varepsilon$ from the variational distribution. To sample $\varepsilon$, we sample $u\sim q^{\textrm{(rep)}}(u)$ and set $\varepsilon=T(u\prm\nu)$. 
To alleviate the computational complexity, we apply the reduce step, sampling a random subset $\Scal\subseteq\{1,\ldots,K\}\diagdown\{k\}$ of outcomes. We thus form a one-sample gradient estimator as 
\begin{equation}\label{eq:gradient_elbo_approx_othermodels}
	\widetilde{\nabla}_{\nu}\Lcal = \nabla_{\varepsilon} \log \widetilde{p}(y,\varepsilon\g\psi) \nabla_{\nu} T(u\prm\nu) + \nabla_{\nu}\mathbb{H}[q(\varepsilon\prm\nu)],
\end{equation}
where $\mathbb{H}[q(\varepsilon\prm\nu)]$ is the entropy of the variational distribution,\footnote{We can estimate the gradient of the entropy when it is not available analytically. Even when it is, the Monte Carlo estimator may have lower variance \citep{Roeder2017}.}
and $\log \widetilde{p}(y,\varepsilon\g\psi) $ is a log joint estimate,
\begin{equation*}
	\log \widetilde{p}(y,\varepsilon\g\psi) = \log\phi(\varepsilon) + \frac{K-1}{|\Scal|}\!\sum_{k^\prime\in\Scal} \log\Phi(\varepsilon + \psi_{k}-\psi_{k^\prime}).
\end{equation*}

In the global step, we estimate the gradient of the \gls{ELBO} with respect to $w$. Following a similar approach, we obtain an unbiased one-sample gradient estimator as
$	\widetilde{\nabla}_{w}\Lcal = \frac{K-1}{|\Scal|}\sum_{k^\prime\in\Scal} \nabla_{w} \log\Phi(\varepsilon + \psi_{k}-\psi_{k^\prime})$.


\newlength{\textfloatsepsavebis}
\setlength{\textfloatsepsavebis}{\textfloatsep}
\setlength{\textfloatsep}{0.10in} 

\begin{algorithm}[t]
	\caption{General \acrshort{AR} for classification}
	\label{alg:augmented_general_vi}
	\begin{algorithmic}
		\STATE {\bfseries Input:} data $(x_n,y_n)$, minibatch sizes $|\Bcal|$ and $|\Scal|$
		\STATE {\bfseries Output:} weights $w=\{w_k\}_{k=1}^K$
		\STATE Initialize all weights and local variational parameters
		\FOR{iteration $t=1,2,\ldots,$}
			\STATE \verb|# Sample minibatches:|
			\STATE Sample a minibatch of data, $\Bcal\subseteq\{1,\ldots,N\}$
			\FOR{$n\in\Bcal$}
				\STATE Sample a set of labels, $\Scal_n\subseteq\{1,\ldots,K\}\diagdown\{y_n\}$
			\ENDFOR
			\STATE \verb|# Local step (E step):|
			\FOR{$n\in\Bcal$}
				\STATE Sample auxiliary variable $u_n\sim q^{\textrm{(rep)}}(u_n)$
				\STATE Transform auxiliary variable, $\varepsilon_n=T(u_n\prm\nu_n)$
				\STATE Estimate the gradient $\widetilde{\nabla}_{\nu_n}\!\Lcal^{(n)}$ (\Cref{eq:gradient_elbo_approx_othermodels})
				\STATE Update variational param., $\nu_n \leftarrow \nu_n + \alpha^{(t)} \widetilde{\nabla}_{\nu_n}\Lcal^{(n)}$
			\ENDFOR
			\STATE \verb|# Global step (M step):|
			\STATE Sample $\varepsilon_n\sim q(\varepsilon_n\prm\nu_n)$ for all $n\in\Bcal$
			\STATE Set $g \!=\! \frac{N}{|\Bcal|}\frac{K-1}{|\Scal|} \substack{\sum \\ n\in \Bcal} \; \substack{\sum \\ k^\prime\in\Scal_n} \!\! \nabla_{w} \! \log\Phi(\varepsilon_n\!+\psi_{ny_n}\!-\psi_{nk^\prime})$
			\STATE Gradient step on the weights, $w \leftarrow w + \rho^{(t)}g$
		\ENDFOR
	\end{algorithmic}
\end{algorithm}

\setlength{\textfloatsep}{\textfloatsepsavebis}

\Cref{alg:augmented_general_vi} summarizes the procedure to efficiently run maximum likelihood on a classification problem. We subsample observations and classes at each iteration.

Finally, note that we can perform posterior inference on the parameters $w$ by positing a variational distribution $q(w)$ and taking gradient steps with respect to the variational parameters of $q(w)$ in the global step. In this case, the reparameterization trick is needed in both the local and global step to obtain Monte Carlo estimates of the gradient.

We now particularize \gls{AR} for the multinomial probit and multinomial logistic models.

\parhead{\acrshort{AR} for multinomial probit.}
Consider a standard Gaussian distribution over the error terms,
\begin{equation*}
	\phi_{\textrm{probit}}(\varepsilon) = \frac{1}{\sqrt{2\pi}}e^{-\frac{1}{2}\varepsilon^2},  
	\quad
	\Phi_{\textrm{probit}}(\varepsilon)=\int_{-\infty}^{\varepsilon} \phi_{\textrm{probit}}(\tau) d\tau.
\end{equation*}
\gls{AR} chooses a Gaussian variational distribution $q_{\textrm{probit}}(\varepsilon\prm \nu)=\Ncal(\varepsilon\prm \mu,\sigma^2)$ and fits the variational parameters $\nu=[\mu,\; \sigma]^\top$. The Gaussian is reparameterizable in terms of a standard Gaussian, i.e., $q_{\textrm{probit}}^{\textrm{(rep)}}(u)=\Ncal(u\prm 0,1)$. The transformation is $\varepsilon=T(u\prm \nu)=\mu+\sigma u$. Thus, the gradients in \Cref{eq:gradient_elbo_approx_othermodels} are $\nabla_{\nu} T(u\prm\nu) = [1,\; u]^\top$ and $\nabla_{\nu}\mathbb{H}[q_{\textrm{probit}}(\varepsilon\prm\nu)]=[0, \; 1/\sigma]^\top$.

\parhead{\acrshort{AR} for multinomial logistic.}
Consider now a standard logistic distribution over the errors,
\begin{equation*}
	\phi_{\textrm{logistic}}(\varepsilon) = \sigma(\varepsilon)\sigma(-\varepsilon), \qquad \Phi_{\textrm{logistic}}(\varepsilon)=\sigma(\varepsilon),
\end{equation*}
where $\sigma(\varepsilon)=\frac{1}{1+e^{-\varepsilon}}$ is the sigmoid function. (The logistic distribution has heavier tails than the Gaussian.) Under this model, the \gls{ELBO} in \Cref{eq:elbo} takes the form
\begin{equation*}\label{eq:elbo_logistic}
	\Lcal_{\textrm{logistic}}\! =\! \mathbb{E}_{q(\varepsilon)}\!\Big[ \! \log\frac{\sigma(\varepsilon) \sigma(-\varepsilon)}{q(\varepsilon)} + \!\! \sum_{k^\prime\neq k}\! \log\sigma(\varepsilon + \psi_{k}-\psi_{k^\prime}) \Big].
\end{equation*}
Note the close resemblance between this expression and the \gls{OVE} bound of \citet{Titsias2016}, 
\begin{equation}\label{eq:ove_bound}
	\Lcal_{\textrm{OVE}} = \sum_{k^\prime\neq k} \log\sigma(\psi_{k}-\psi_{k^\prime}).
\end{equation}
However, while the former is a bound on the multinomial logistic model, the \gls{OVE} is a bound on the softmax.

\gls{AR} sets $q_{\textrm{logistic}}(\varepsilon\prm \nu)=\frac{1}{\beta} \sigma\Big(\frac{\varepsilon-\mu}{\beta}\Big)\sigma\Big(-\frac{\varepsilon-\mu}{\beta}\Big)$, a logistic distribution. The variational parameters are $\nu=[\mu,\;\beta]^\top$. The logistic distribution is reparameterizable, with $q_{\textrm{logistic}}^{\textrm{(rep)}}(u)=\sigma(u)\sigma(-u)$ and transformation $\varepsilon=T(u\prm \nu)=\mu+\beta u$. The gradient of the entropy in \Cref{eq:gradient_elbo_approx_othermodels} is $\nabla_{\nu}\mathbb{H}[q_{\textrm{logistic}}(\varepsilon\prm\nu)]=[0, \; 1/\beta]^\top$.

\section{Experiments}
\label{sec:experiments}
\glsresetall

We showcase \gls{AR} on a linear classification task. Our goal is to assess the predictive performance of \gls{AR} in this classification
task, to assess the quality of the marginal bound of the data, and to compare its
complexity\footnote{We focus on runtime cost. \acrshort{AR} requires $\Ocal(N)$ memory storage capacity due to the local variational parameters.}
with existing approaches.

We run \gls{AR} for three different models of categorical distributions
(softmax, multinomial probit, and multinomial
logistic).\footnote{Code for \acrshort{AR} is available at \url{https://github.com/franrruiz/augment-reduce}.}
For the softmax model, we compare \gls{AR} against the \gls{OVE} bound \citep{Titsias2016}.
Just like \gls{AR}, \gls{OVE} is a 
rigorous lower bound on the marginal likelihood. It can also run on a single machine,\footnote{\gls{AR} is amenable to an embarrassingly parallel algorithm, but we focus on single-core procedures.}
and it has been shown to outperform other approaches.

For softmax, \gls{AR} runs nearly as fast as \gls{OVE} but 
has better predictive performance and provides a tighter bound on the marginal likelihood than \gls{OVE}. 
On two small datasets, the \gls{AR} bound closely reaches the marginal likelihood of exact softmax maximum likelihood estimation.

We now describe the experimental settings.
In \Cref{sec:experiment_toy}, we analyze synthetic data and $K=10^4$ classes.
In \Cref{sec:experiment_real}, we analyze five real datasets.

\parhead{Experimental setup.} We consider linear classification, where the mean utilities are $\psi_{nk}=w_k^\top x_n + w_k^{(0)}$. We fit the model parameters (weights and biases) via maximum likelihood estimation, using stochastic gradient ascent. We initialize the weights and biases randomly, drawing from a Gaussian distribution with zero mean and standard deviation $0.1$ ($0.001$ for the biases). For each experiment, we use the same initialization across all methods.

\Cref{alg:augmented_sm_vi,alg:augmented_general_vi} require setting a step size schedule for $\rho^{(t)}$.
We use the adaptive step size sequence proposed by \citet{Kucukelbir2017}, which combines \textsc{rmsprop} \citep{Tieleman2012} and Adagrad \citep{Duchi2011}. We set the step size using the default parameters, i.e.,
\begin{equation*} 
	\begin{split}
		& \rho^{(t)} = \rho_0 \times t^{-1/2+10^{-16}}\times \left(1+\sqrt{s^{(t)}} \right)^{-1}, \\
		& s^{(t)} = 0.1 (g^{(t)})^2 + 0.9 s^{(t-1)}.
	\end{split}
\end{equation*}
We set $\rho_0=0.02$ and we additionally decrease $\rho_0$ by a factor of $0.9$ every $2000$ iterations. We use the same step size sequence for \gls{OVE}.

We set the step size $\alpha^{(t)}$ in \Cref{alg:augmented_sm_vi} as $\alpha^{(t)}=(1+t)^{-0.9}$, the default values suggested by \citet{Hoffman2013}. For the step size $\alpha^{(t)}$ in \Cref{alg:augmented_general_vi}, we set $\alpha^{(t)}=0.01(1+t)^{-0.9}$. For the multinomial logit and multinomial probit \gls{AR}, we parameterize the variational distributions in terms of their means $\mu$ and their unconstrained scale parameter $\gamma$, such that the scale parameter is $\log(1+\exp(\gamma))$.

\subsection{Synthetic Dataset}\label{sec:experiment_toy}

We mimic the toy experiment of \citet{Titsias2016} to assess how well \gls{AR} estimates the categorical probabilities. We generate a dataset with $10^4$ classes and $N=3\times 10^5$ observations, each assigned label $k$ with probability $p_k\propto \widetilde{p}_k^2$, where each $\widetilde{p}_k$ is randomly generated from a uniform distribution in $[0,1]$. After generating the data, we have $K=9{,}035$ effective classes (thus we use this value for $K$). In this simple setting, there are no observed covariates $x_n$.

We estimate the probabilities $p_k$ via maximum likelihood on the biases $w_k^{(0)}$. We posit a softmax model, and we apply both the \gls{VEM} in \Cref{sec:softmax_augment} and the \gls{OVE} bound. For both approaches, we choose a minibatch size of $|\Bcal|=500$ observations and $|\Scal|=100$ classes, and we run $5\times 10^5$ iterations.


\begin{table*}[t]
 	\centering
 	\small
 	\caption{Statistics and experimental settings of the considered datasets. $N_{\textrm{train}}$ and $N_{\textrm{test}}$ are the number of training and test data points. The number of classes is the resulting value after the preprocessing step (see text). The minibatch sizes correspond to $|\Bcal|$ and $|\Scal|$, respectively.\label{tab:data_stat}}
 	\begin{tabular}{c} \toprule
 		 dataset  \\ \midrule
 		 MNIST     \\ 
 		 Bibtex \\
 		 Omniglot     \\
 		 EURLex-4K \\
 		 AmazonCat-13K \\ \bottomrule
 	\end{tabular}
 	\hspace{1pt}
 	\begin{tabular}{cccc} \toprule
 		    $N_{\textrm{train}}$  &	$N_{\textrm{test}}$ &	 covariates         & classes  \\ \midrule
 			$60,000$    &	$10,000$   &	$784$       & $10$	         \\ 
 		    $4,880$  &   $2,413$  &    $1,836$   &   $148$       \\
 			$25,968$            &	$6,492$            &	$784$           & $1,623$	 \\
 		    $15,539$   &   $3,809$      &    $5,000$     &  $896$        \\
 		    $1,186,239$ &    $306,782$ &    $203,882$ &  $2,919$      \\ \bottomrule
 	\end{tabular}
 	\hspace{1pt}
 	\begin{tabular}{ccc} \toprule
 		minibatch (obs.)  &	minibatch (classes) & iterations   \\ \midrule
 		$500$ & $1$  &  $35,000$	         \\ 
 		$488$ & $20$ &  $5,000$ \\
 		$541$ & $50$ &  $45,000$	         \\
 		$379$ & $50$ &  $100,000$           \\
 		$1,987$ & $60$ &  $5,970$          \\ \bottomrule
 	\end{tabular}
	\vspace*{-10pt}
\end{table*}


\begin{figure}[tb]
	\centering
	\includegraphics[width=0.24\textwidth]{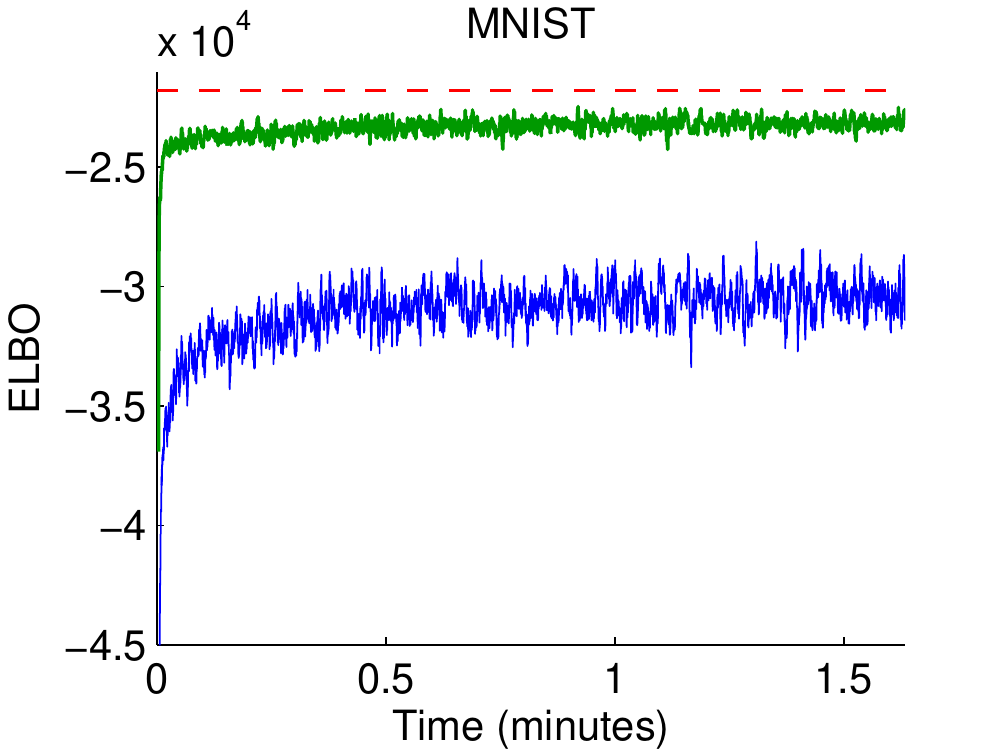} \hspace*{-4pt}
	\includegraphics[width=0.24\textwidth]{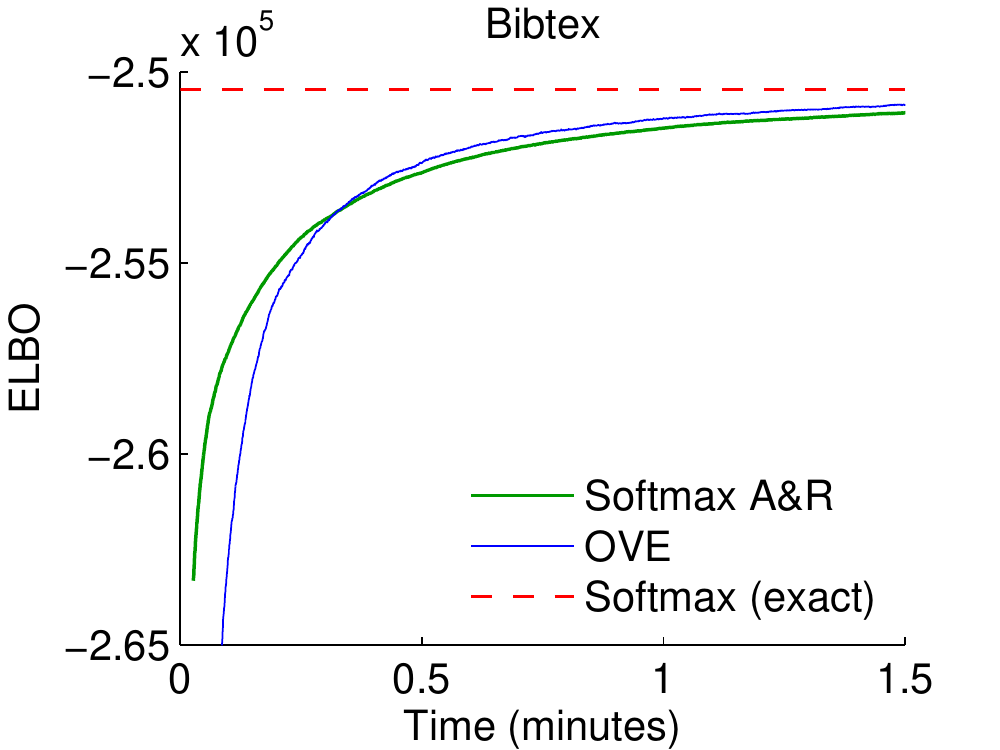} \\ \vspace*{2pt}
	\includegraphics[width=0.24\textwidth]{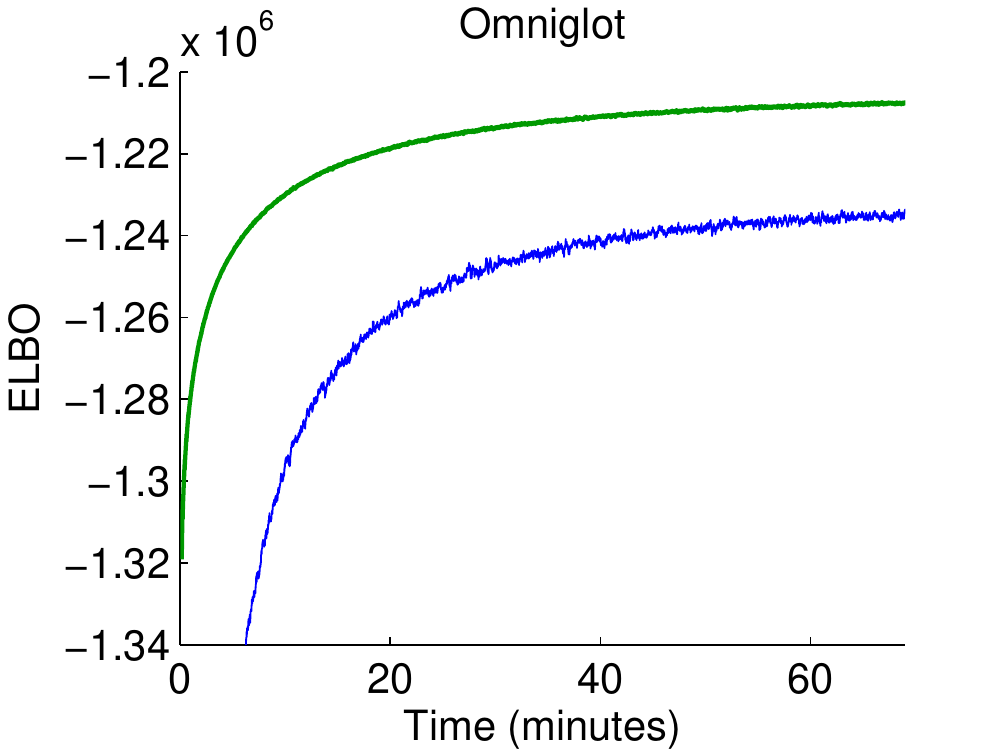} \hspace*{-4pt}
	\includegraphics[width=0.24\textwidth]{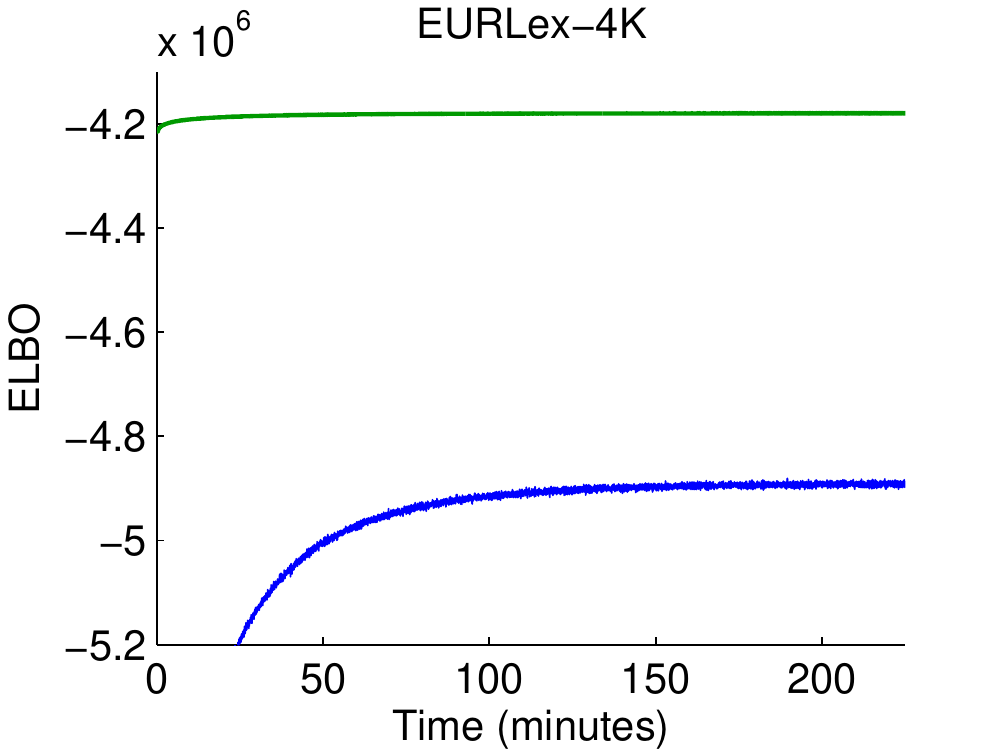} \\ \vspace*{2pt}
	\includegraphics[width=0.24\textwidth]{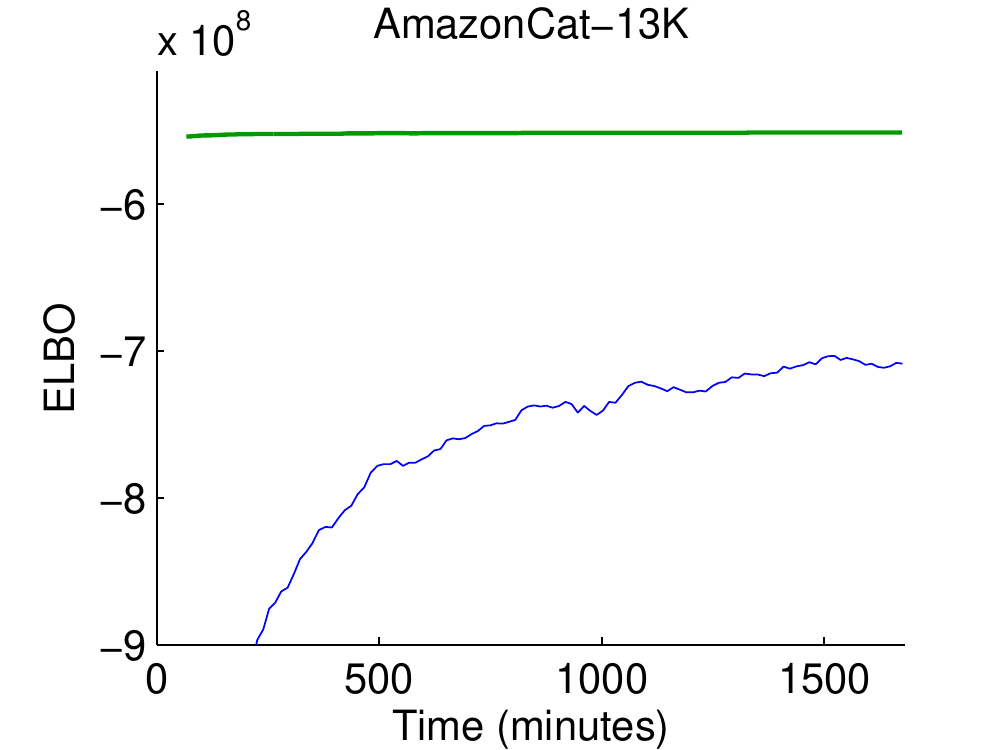}
    \vspace*{-10pt}
	\caption{Evolution of the \acrshort{ELBO} as a function of wall-clock time. The softmax \acrshort{AR} method provides a tighter bound than \acrshort{OVE} \citep{Titsias2016} for almost all the considered datasets.\label{fig:elbo_evolution}}
    \vspace*{-14pt}
\end{figure}

We run each approach on one \textsc{cpu} core. On average, the wall-clock time per epoch (one epoch takes $N/|\Bcal|=600$ iterations) is $0.196$ minutes for softmax \gls{AR} and $0.189$ minutes for \gls{OVE}. \gls{AR} is slightly slower because of the local step that \gls{OVE} does not require; however, the bound on the marginal log likelihood is tighter (by orders of magnitude) for \gls{AR} than for \gls{OVE} ($-2.62\times 10^6$ and $-1.40\times 10^9$, respectively). The estimated probabilities are similar for both methods: the average absolute error is $3.00\times 10^{-6}$ for \gls{AR} and $3.65\times 10^{-6}$ for \gls{OVE}; the difference is not statistically significant.

\subsection{Real Datasets}\label{sec:experiment_real}

We now turn to real datasets.
We consider MNIST and Bibtex \citep{Katakis2008,Prabhu2014},
where we can compare against the exact softmax. We also analyze 
Omniglot \citep{Lake2015}, EURLex-4K \citep{Mencia2008,Bhatia2015},
and AmazonCat-13K \citep{McAuley2013}.%
\footnote{MNIST is available at \url{http://yann.lecun.com/exdb/mnist}. 
Omniglot can be found at \url{https://github.com/brendenlake/omniglot}.
Bibtex, EURLex-4K, and AmazonCat-13K are available 
at \url{http://manikvarma.org/downloads/XC/XMLRepository.html}.
} 
\Cref{tab:data_stat} gives information about the structure of these datasets. 

We run each method for a fixed number of iterations. We set the minibatch sizes $|\Bcal|$ and $|\Scal|$ beforehand. The specific values for each dataset are also in \Cref{tab:data_stat}.


\begin{table*}[t]
 	\centering
 	\small
 	\caption{Average time per epoch for each method and dataset. Softmax \acrshort{AR} (\Cref{sec:softmax_augment}) is almost as fast as \acrshort{OVE}. The \acrshort{AR} approaches in \Cref{sec:other_augment} take longer because they require some additional computations, but they are still competitive.\label{tab:running_time}}
 	\begin{tabular}{c} \toprule
 		 \\
 		 dataset  \\ \midrule
 		 MNIST     \\ 
 		 Bibtex \\
 		 Omniglot     \\
 		 EURLex-4K \\
 		 AmazonCat-13K \\ \bottomrule
 	\end{tabular}
 	\hspace{1pt}
 	\begin{tabular}{c} \toprule
 		\\
 		\acrshort{OVE} \citep{Titsias2016} \\ \midrule
 			$0.336$ s	    \\ 
 		    $0.181$ s       \\
 			$4.47$ s        \\
 		    $5.54$ s        \\
 		    $2.80$ h    \\ \bottomrule
 	\end{tabular}
 	\hspace{1pt}
 	\begin{tabular}{ccc} \toprule
 		\multicolumn{3}{c}{\acrshort{AR} [this paper]} \\
 		softmax & multi.\ probit & multi.\ logistic   \\ \midrule
 		$0.337$ s   &  $0.431$ s  &  $0.511$ s    \\ 
 		$0.188$ s   & $0.244$ s & $0.246$ s \\
 		$4.65$ s   &  $5.63$ s  &  $5.57$ s    \\
 		$5.65$ s   &  $6.46$ s  &  $6.23$ s           \\
 		$2.80$ h   &  $2.82$ h  &  $2.91$ h     \\ \bottomrule
 	\end{tabular}
	\vspace*{-10pt}
\end{table*}


\begin{table*}[!hbpt]
 	\centering
 	\small
 	\caption{Test log likelihood and accuracy for each method and dataset. The table on the left compares the approaches based on the softmax. Softmax \acrshort{AR} outperforms \gls{OVE} in four out of the five datasets. The two tables on the right show the performance of other models (multinomial probit and multinomial logistic), for which \acrshort{AR} is also competitive.\label{tab:pred_llh_acc}}
 	\begin{tabular}{c} \toprule
 		 \\
 		 \\
 		 dataset  \\ \midrule
 		 MNIST     \\ 
 		 Bibtex \\
 		 Omniglot     \\
 		 EURLex-4K \\
 		 AmazonCat-13K \\ \bottomrule
 	\end{tabular}
 	\hspace{1pt}
 	\begin{tabular}{cc|cc|cc} \toprule
 		 \multicolumn{6}{c}{softmax model}   \\
 		 \multicolumn{2}{c|}{exact} & \multicolumn{2}{c|}{\acrshort{OVE} \citep{Titsias2016}}  & \multicolumn{2}{c}{\acrshort{AR} [this paper]}   \\
 		 log lik & acc & log lik & acc & log lik & acc \\ \midrule
 		 $-0.261$ & $0.927$ & $-0.276$ &  $0.919$  &  $\mathbf{-0.271}$   &  $\mathbf{0.924}$     \\ 
 		 $-3.188$ & $0.361$ &  $-3.300$  &  $0.352$ &  $\mathbf{-3.036}$   &  $\mathbf{0.361}$ \\
 		 $-$ & $-$ &  $-5.667$ &  $0.179$  &  $\mathbf{-5.171}$   &  $\mathbf{0.201}$     \\
 		 $-$ & $-$ &  $\mathbf{-4.241}$ &  $\mathbf{0.247}$  &  $-4.593$   &  $0.207$     \\
 		 $-$ & $-$ &  $-3.880$ & $0.388$ & $\mathbf{-3.795}$ & $\mathbf{0.420}$ \\ \bottomrule
 	\end{tabular}
 	\hspace{1pt}
 	\begin{tabular}{cc} \toprule
 		\multicolumn{2}{c}{multi.\ probit}  \\
 		\multicolumn{2}{c}{\acrshort{AR} [this paper]} \\
 		log lik & acc \\ \midrule
 		$-0.302$ &  $0.918$      \\ 
 		$-4.184$ &  $0.346$      \\
 		$-7.350$ &  $0.178$      \\
 		$-4.193$ &  $0.263$      \\
 		$-3.593$ &  $0.411$ \\ \bottomrule
 	\end{tabular}
 	\hspace{1pt}
 	\begin{tabular}{cc} \toprule
 		\multicolumn{2}{c}{multi.\ logistic}  \\
 		\multicolumn{2}{c}{\acrshort{AR} [this paper]} \\
 		log lik & acc \\ \midrule
 		$-0.287$   &  $0.917$     \\ 
 		$-3.151$ &  $0.353$ \\
 		$-5.395$   &  $0.184$     \\
 		$-4.299$   &  $0.226$     \\
 		$-4.081$   &  $0.350$ \\ \bottomrule
 	\end{tabular}
	\vspace*{-8pt}
\end{table*}

\parhead{Data preprocessing.} For MNIST, we divide the pixel values by $255$ so that the maximum value is one. For Omniglot, following other works in the literature \citep[e.g.,][]{Burda2016}, we resize the images to $28\times 28$ pixels. For EURLex-4K and AmazonCat-13K, we normalize the covariates dividing by their maximum value.

Bibtex, EURLex-4K, and AmazonCat-13K are multi-class datasets, i.e., each observation may be assigned more than one label. Following \citet{Titsias2016}, we keep only the first non-zero label for each data point. See \Cref{tab:data_stat} for the resulting number of classes in each case.

\parhead{Evaluation.} For the softmax, we compare \gls{AR} against the \gls{OVE}
bound.\footnote{We also implemented the approach of \citet{Botev2017}, but we do not report the results because it did not outperform \gls{OVE} in terms of test log-likelihood on four out of the five considered datasets. On the fifth dataset, softmax \gls{AR} was still superior.}
We also compare against the exact softmax on MNIST and Bibtex, where the number of classes is small. 
For the multinomial probit and multinomial logistic models, we also report the predictive performance of \gls{AR}.

We evaluate performance with test log likelihood and accuracy. The accuracy is the fraction of correctly classified instances, assuming that we assign the most likely label (i.e., the one with the highest mean utility). To compute the test log likelihood, we use \Cref{eq:softmax} for the softmax and \Cref{eq:girolami_rogers} for the multinomial probit and multinomial logistic models. We approximate the integral in \Cref{eq:girolami_rogers} with $1{,}000$ samples using importance sampling (we use a Gaussian distribution with mean $5$ and standard deviation $5$ as a proposal). 

\parhead{Results.}
\Cref{tab:running_time} shows the wall-clock time per epoch for each method and dataset. In general, softmax \gls{AR} is almost as fast as \gls{OVE} because the extra local step can be performed efficiently without additional expensive operations. It requires to evaluate exponential functions that can be reused in the global step. Multinomial probit \gls{AR} and multinomial logistic \gls{AR} are slightly slower because of the local step, but they are still competitive.

For the five datasets,
\Cref{fig:elbo_evolution} shows the evolution of the \gls{ELBO} as a function of wall-clock time for the softmax \gls{AR} (\Cref{eq:elbo_softmax}), compared to the \gls{OVE} (\Cref{eq:ove_bound}). For easier visualization, we plot a smoothed version of the bounds after applying a moving average window of size $100$. (For AmazonCat-13K, we only compute the \gls{ELBO} every $50$ iterations and we use a window of size $5$.)
Softmax \gls{AR} provides a significantly tighter bound for most datasets (except for Bibtex, where the \gls{ELBO} of \gls{AR} is close to the \gls{OVE} bound). For MNIST and Bibtex, we also plot the marginal likelihood obtained after running maximum likelihood estimation on the exact softmax model. The \gls{ELBO} of \gls{AR} nearly achieves this value.

Finally, \Cref{tab:pred_llh_acc} shows the predictive performance for all methods across all datasets. We report test log likelihood and accuracy. Softmax \gls{AR} outperforms \gls{OVE} in both metrics on all but one dataset (except EURLex-4K). Although our goal is not to compare performance across different models, for completeness \Cref{tab:pred_llh_acc} also shows the predictive performance of multinomial probit \gls{AR} and multinomial logistic \gls{AR}. In general, softmax \gls{AR} provides the highest test log likelihood, but multinomial probit \gls{AR} outperforms all other methods in EURLex-4K and AmazonCat-13K. Additionally, multinomial logistic \gls{AR} presents better predictive performance than \gls{OVE} on Omniglot and Bibtex.

\section{Conclusion}
\label{sec:conclusion}
\glsresetall

We have introduced \gls{AR}, a scalable method to
fit models involving categorical distributions. 
\gls{AR} is general and applicable to many models, including the softmax and the 
multinomial probit. 
On classification tasks, we found that 
\gls{AR} outperforms state-of-the art algorithms with little extra computational cost. 



\section*{Acknowledgements}
This work was supported by ONR N00014-15-1-2209, ONR 133691-5102004, NIH 5100481-5500001084, NSF CCF-1740833, the Alfred P.\ Sloan Foundation, the John Simon Guggenheim Foundation, Facebook, Amazon, and IBM.
Francisco J.\ R.\ Ruiz is supported by the EU Horizon 2020 programme (Marie Sk\l{}odowska-Curie Individual Fellowship, grant agreement 706760).
We also thank Victor Elvira and Pablo Moreno for their comments and help.

\bibliography{augmented_softmax}
\bibliographystyle{icml2018}

\end{document}